# Finding fake reviews in e-commerce platforms by using hybrid algorithms

Mathivanan Periasamy (mathivananperiasamy@gmail.com), Rohith Mahadevan(mailrohithmahadevan@gmail.com), Bagiya Lakshmi S(bagiyalakshmi59@gmail.com),
Raja CSP Raman (raja.csp@gmail.com), Hasan Kumar S (hasankumarsenthil@gmail.com),
Jasper Jessiman (jessimanjasper2.123@gmail.com)

**Abstract**

Sentiment analysis, a vital component in natural language processing, plays a crucial role in understanding the underlying emotions and opinions expressed in textual data. In this paper, we propose an innovative ensemble approach for sentiment analysis for finding fake reviews that amalgamate the predictive capabilities of Support Vector Machine (SVM), K-Nearest Neighbors (KNN), and Decision Tree classifiers. Our ensemble architecture strategically combines these diverse models to capitalize on their strengths while mitigating inherent weaknesses, thereby achieving superior accuracy and robustness in fake review prediction. By combining all the models of our classifiers, the predictive performance is boosted and it also fosters adaptability to varied linguistic patterns and nuances present in real-world datasets. The metrics accounted for on fake reviews demonstrate the efficacy and competitiveness of the proposed ensemble method against traditional single-model approaches. Our findings underscore the potential of ensemble techniques in advancing the state-of-the-art in finding fake reviews using hybrid algorithms, with implications for various applications in different social media and e-platforms to find the best reviews and neglect the fake ones, eliminating puffery and bluffs. Github: https://github.com/tactlabs/fake-review-hybrid-algorithm

## 1 Introduction

The burgeoning field of Natural Language Processing (NLP) has been pivotal in deciphering the vast expanses of textual data generated by digital platforms. Among its myriad applications, sentiment analysis emerges as a critical tool for discerning the nuanced emotions and opinions embedded within the text. This capability is not only fundamental for understanding user sentiments across social media and e-commerce platforms but also instrumental in identifying deceptive practices such as the dissemination of fake reviews. The integrity of online reviews is paramount, as they significantly influence consumer behavior and perception of products and services. Consequently, the detection and filtration of fake reviews are essential to maintaining the authenticity and trustworthiness of digital platforms.

In the digital age, where user-generated content has become a cornerstone of consumer decision-making, the proliferation of fake reviews threatens to undermine the credibility of online platforms. Traditional sentiment analysis techniques, while effective to a certain extent, often fall short when faced with the sophistication and variability of fake reviews. To address this challenge, our research introduces an innovative ensemble approach that synergizes the strengths of Support Vector Machine (SVM), K-Nearest Neighbors (KNN), and Decision Tree classifiers. This paper proposes a comprehensive framework for sentiment analysis, specifically tailored to the identification and analysis of fake reviews. Our ensemble method is designed to leverage the predictive capabilities of these diverse models, enhancing accuracy and robustness in detecting counterfeit sentiments.

The research encapsulates a series of methodical steps encompassing data preparation, preprocessing, feature extraction, and the deployment of various machine learning models, culminating in the ensemble approach. Through meticulous data curation and innovative algorithmic strategies, our study aims to set a new benchmark in sentiment analysis for fake review detection. The proposed ensemble architecture not only signifies a leap in predictive performance but also demonstrates adaptability to the complex linguistic patterns and nuances characteristic of fake reviews. By addressing the limitations of traditional single-model approaches, our research underscores the potential of hybrid algorithms in navigating the challenges posed by fake reviews on social media and e-commerce platforms.

This paper is structured to guide the reader through the intricacies of our methodology, from the initial stages of data preparation and preprocessing to the intricate processes of model selection and feature extraction. Our findings illuminate the efficacy of the ensemble approach in discerning genuine from deceptive reviews, offering a beacon of hope for entities vested in the integrity of online discourse. As we traverse through the realms of machine learning and sentiment analysis, this study aspires to contribute valuably to the ongoing efforts to safeguard the digital ecosystem against the scourge of misinformation and deceit.

## 2 Literature Review

In the growing realm of e-commerce, dealing with the surge of fake reviews poses a significant challenge, for both customers and companies. To address this issue researchers are exploring methods, such as machine learning, deep learning, and hybrid approaches. They aim to create systems to identify and counter deceptive practices. These studies dive into the details of review authenticity by using innovative techniques to navigate through the vast ocean of online reviews.

Choi and colleagues [1], and Han and associates [2] introduce machine learning and Explainable Knowledge Integrated Sequence Model (EKI SM) respectively focusing on boosting trust in online shopping through review categorization. Meanwhile, Elmogy and team [3] highlight behavioral traits in their machine-learning technique, for detecting fake reviews.

In the field of deep learning, Alsubari and team [4], and Qayyum and colleagues [5] delve into learning techniques such as CNN BiLSTM and FRD LSTM showcasing accuracy, in detecting fraudulent reviews. Zhang and associates [6] suggest a framework that takes into account both behavioral and textual cues while Mohawesh and team [7] introduce a multi-view deep learning model that integrates various perspectives.

Salminen and colleagues [8] discover that machine classifiers outperform human classifiers, whereas Lu and associates [9] present BSTC, a Fake Review Detection Model that outshines existing methods. Bathla and team [10] concentrate on aspect extraction and analysis through deep learning techniques achieving superior performance.

Hajek and colleagues [11] combined word embeddings with sentiment analysis demonstrating superior performance in fake review detection. Deshai and Rao [12], and Duma and team [13] explored hybrid models that merge CNN with adaptive particle swarm optimization and deep hybrid models respectively, both yielding promising outcomes.

Mohawesh and colleagues [14] delved into concept drift analysis emphasizing its influence on fake review detection. Abdulqader and team [15] introduced a Unified Detection Model based on psychology while Vidanagama and team [16] utilized ontology-based sentiment analysis to enhance detection accuracy.

Hajek and team [17] emphasized aspect-based sentiment analysis, Shunxiang and colleagues [18] introduced a PU learning-based technique, and Jing Yu and colleagues [19] proposed a Supervised Fake Reviews Detection method using AspamGAN—each showcasing advancements in detection precision. Finally, Luo and team [20] devised a probability-based strategy to enhance the accuracy of detecting fake reviews on platforms.

While our ensemble approach demonstrates high effectiveness in detecting fake reviews within static datasets, the dynamic nature of online platforms necessitates ongoing adaptation to counter new deceptive tactics. Furthermore, our research, while focused on textual analysis, acknowledges the potential value in incorporating non-textual features such as reviewer behavior patterns, temporal variations, and contextual cues, which are often overlooked in traditional sentiment analysis. Additionally, the importance of testing fake review detection models across a variety of platforms is recognized to ensure the robustness and generalizability of our proposed solutions. Many studies are constrained by their reliance on a narrow range of datasets, which may introduce biases and restrict the broader applicability of their findings. Our research aims to address these gaps by advocating for a more holistic and adaptable approach to fake review detection.

## 3 Methodology and Component Explanation

### 3.1. Data Preparation:
Data preparation is crucial for any machine learning task as it lays the foundation for model training. In this step, we initially have data in CSV format containing columns such as URL, Reviewers' Rating, Review, and Collected By. Additionally, we have separate CSV files for different product reviews. We concatenate all these CSV files into one comprehensive dataset.

Next, we manually label the data based on specific criteria such as Extreme Negative or Positive Emotion, Personal Stories and Details, Excellent or Poor Grammar, Excessive Humor, and a Focus on Irrelevant Details. This labeling helps in supervised learning where the model learns from labeled examples. Each review is assigned a binary label indicating whether it is considered Fake/Useless (0) or Good/Useful (1).

### 3.2. Preprocessing steps:

Preprocessing the data is essential for cleaning and transforming it into a format suitable for analysis and model training.
**Handling Null Values**: We remove rows with null values based on the sentiment column to ensure data quality. **Lowercasing**: Convert all text to lowercase to ensure uniformity and avoid redundancy during feature extraction. **Removing Punctuations**: Punctuation marks do not add much value in sentiment analysis and are removed to simplify the text. **Tokenization**: Splitting the text into individual tokens (words) to prepare it for further processing. **Stopword Removal**: Eliminating common words like "and", "the", etc., which do not carry significant meaning in sentiment analysis. **Stemming**: Reducing words to their root form to normalize variations of the same word (e.g., "running" to "run"). **Lemma Building**: Similar to stemming but retains the actual meaning of words by converting them into their base or dictionary form. **Removing Emojis**: Emojis can convey sentiment but might not be processed effectively by all models, hence they are removed for consistency.

### 3.3. Feature Extraction:

Feature extraction involves converting raw text into numerical features that machine learning models can understand. **Word2Vec**: Utilizing the Word2Vec algorithm from the gensim library to convert words into dense vectors, capturing semantic similarities between words. **BERT**: Employing the pre-trained BERT (Bidirectional Encoder Representations from Transformers) model, specifically the bert-base-uncased variant, to generate contextualized word embeddings, which capture intricate linguistic patterns. **TF-IDF**: Using techniques such as CountVectorizer and TfidfVectorizer to convert text into numerical vectors based on the term frequency-inverse document frequency (TF-IDF) principle, which highlights the importance of words in a document relative to the entire corpus.

### 3.4. Model:

Various machine learning models are employed for sentiment analysis:
**Decision Tree**: A decision tree classifier with specific hyperparameters (random_state=9, max_depth=5) is used to build a predictive model based on learned decision rules.
**Random Forest**: An ensemble learning method, comprising multiple decision trees, is utilized with the same hyperparameters as the decision tree model.
**Logistic Regression**: A linear model suitable for binary classification tasks, configured with random_state=9 and max_iter=1000 for convergence.

$$sigma(z) = \frac{1}{1 + e^{-z}}$$

**KNN** (K-Nearest Neighbors): A non-parametric method that classifies instances based on the majority class of their k nearest neighbors.

**SVM** (Support Vector Machine): A powerful supervised learning algorithm that constructs hyperplanes in a high-dimensional space to separate data into different classes.

$$f(x) = \text{sign}\left(\sum_{i=1}^{n} \alpha_i y_i K(x_i, x) + b\right)$$

**Naive Bayes**: A probabilistic classifier based on Bayes' theorem with the assumption of independence between features.

$$P(y|x_1, x_2, \ldots, x_n) = \frac{P(y) \times P(x_1|y) \times P(x_2|y) \times \ldots \times P(x_n|y)}{P(x_1) \times P(x_2) \times \ldots \times P(x_n)}$$

Ensemble (Combining SVM, KNN, and Decision Tree): An ensemble method combining predictions from multiple base models to improve overall accuracy and robustness.

## 4 Proposed Solution

The proposed solution for sentiment analysis involves leveraging an ensemble method that combines Support Vector Machine (SVM), K-Nearest Neighbors (KNN), and Decision Tree classifiers. This ensemble architecture is designed to harness the strengths of each individual model and mitigate their weaknesses, ultimately resulting in enhanced accuracy and robustness in sentiment prediction.

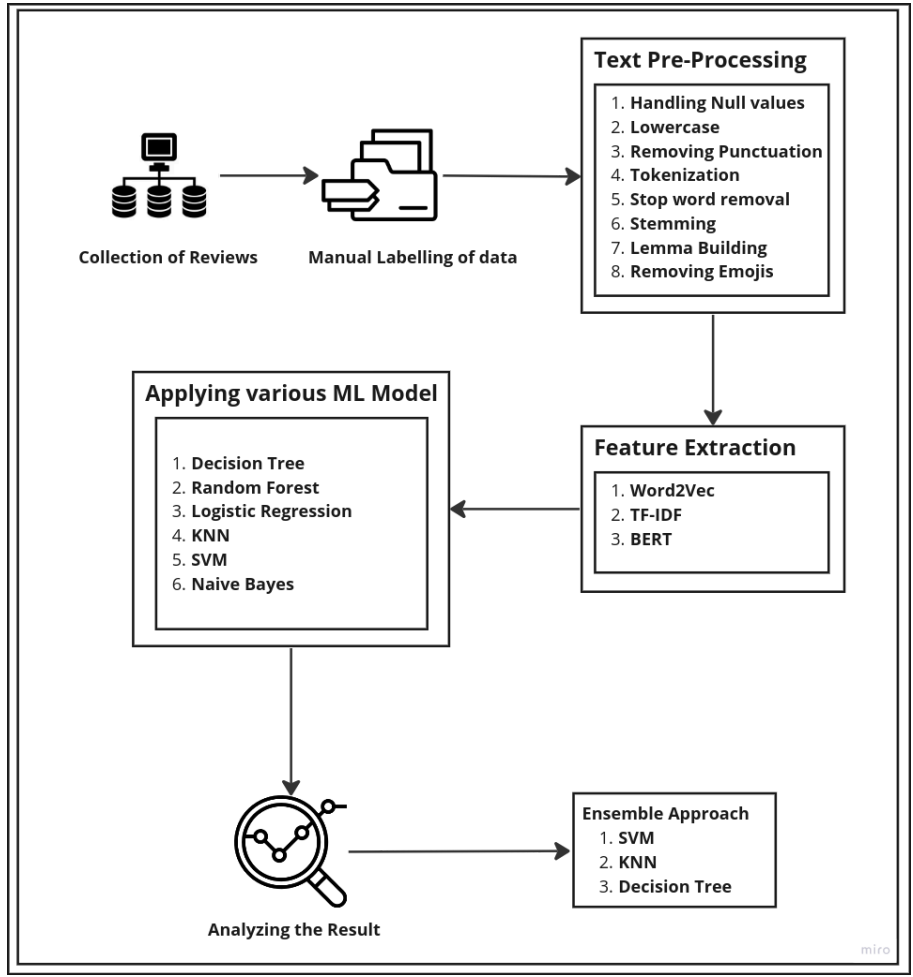

**Key Components:**

SVM (Support Vector Machine):

SVM is a powerful supervised learning algorithm that constructs hyperplanes in a high-dimensional space to separate data into different classes. It is particularly effective in handling complex decision boundaries and is robust to overfitting, making it a valuable component in the ensemble. SVM's ability to handle both linear and non-linear relationships in data adds versatility to the ensemble architecture.

KNN (K-Nearest Neighbors):

KNN is a non-parametric method that classifies instances based on the majority class of their k nearest neighbors. It excels in capturing local patterns and is effective in scenarios where decision boundaries are not well-defined. By incorporating KNN into the ensemble, we leverage its ability to capture subtle nuances in sentiment that may not be captured by other models.

Decision Tree:

Decision trees are intuitive and interpretable models that make predictions by partitioning the feature space into regions based on learned decision rules. They are capable of capturing complex interactions between features and are resilient to outliers and irrelevant attributes.The decision tree classifier adds interpretability to the ensemble, allowing for a deeper understanding of the underlying sentiment patterns.

Ensemble Strategy:

This methodology integrates three distinct machine learning algorithms: the Decision Tree Classifier, Support Vector Machine (SVM), and K-Nearest Neighbors (KNN), each selected for its unique ability to analyze and interpret complex patterns within data. Utilizing the VotingClassifier from the sklearn.ensemble module, our ensemble model employs a 'hard' voting mechanism, where the final sentiment classification of a review is determined by the majority vote across all classifiers.

The configuration of our ensemble model is meticulously chosen to optimize performance and ensure the reproducibility of results. Specifically, the Decision Tree Classifier is set with a max_depth of 5 to prevent overfitting, while both the Decision Tree and SVM are initiated with a random_state of 9, anchoring the stochastic elements of the algorithms for consistent outcomes across different runs. This careful assembly and calibration of the ensemble model underscore its robustness and reliability in identifying fake reviews.

## 5 Results and Analysis

| Model/ Text representation | Random Forest | Decision Tree | Logistic Regression | Ensemble (rf, dt, lr) | SVM | NB | KNN | Ensemble (dt, svm, knn) |
|---|---|---|---|---|---|---|---|---|
| TF-IDF | 0.66 | 0.769 | 0.748 | 0.754 | 0.765 | 0.46 | 0.65 | 0.78 |
| Word2Vec | 0.73 | 0.752 | 0.752 | 0.75 | 0.76 | 0.61 | 0.74 | 0.77 |
| BERT | 0.71 | 0.78 | 0.723 | 0.758 | 0.787 | 0.6811 | **0.782** | **0.801** |

### 5.1 Performance Metrics

In the assessment of our ensemble model's performance, accuracy served as the primary metric, reflecting the model's capability to correctly identify fake reviews.
Our ensemble model, combining SVM, KNN, and Decision Tree classifiers with BERT for feature extraction, achieved a leading accuracy of 80% in detecting fake reviews, setting a new benchmark in sentiment analysis. Notably, the integration of KNN with BERT followed closely, securing a 78% accuracy rate. These results highlight the effectiveness of our approach and the potential of combining machine learning with Large Language Models to improve review authenticity.

## 6 Future Scope

In our research, we've utilized BERT for feature extraction, enhancing our model's ability to identify fake reviews. Looking ahead, we plan to deeply integrate Large Language Models (LLMs) like GPT and BERT's successors to further advance our ensemble approach in sentiment analysis. This integration aims to:

**Enhance Detection Capabilities**: By leveraging the advanced contextual understanding of LLMs, our model will better interpret nuanced sentiments, improving accuracy in spotting fake reviews.
**Enable Continuous Learning**: Incorporating LLMs that learn continuously will allow our model to adapt to new deceptive tactics in real-time, maintaining its effectiveness.
**Expand Multilingual and Cross-domain Use**: We aim to use LLMs' multilingual capabilities to adapt our model for various languages and sectors, broadening its applicability.

## 7 Conclusion

In conclusion, our research introduces a cutting-edge ensemble approach for sentiment analysis, specifically tailored for the detection of fake reviews. By integrating the strengths of SVM, KNN, and Decision Tree classifiers, we have developed a model that showcases superior accuracy and robustness. The incorporation of BERT for feature extraction has already set a high benchmark in understanding complex linguistic patterns. Looking forward, the planned deeper integration of Large Language Models (LLMs) promises to revolutionize our approach, making it more adaptable, efficient, and applicable across various languages and domains. This research not only advances the field of sentiment analysis but also offers practical implications for maintaining the integrity of online review ecosystems. As we continue to refine and expand our methodology, the potential for creating a more authentic digital landscape becomes increasingly attainable.